\documentclass[11pt,a4paper]{article}
\usepackage[hyperref]{emnlp-ijcnlp-2019}
\usepackage{times}
\usepackage{latexsym}
\usepackage{graphicx}
\usepackage{url}
\usepackage{multirow}
\usepackage{latexsym,bm}        %
\usepackage{amsmath,amssymb} 

\usepackage{titlesec}

\aclfinalcopy 

\title{Attention Optimization for Abstractive Document Summarization}

\author{Min Gui\textsuperscript{1}, Junfeng Tian\textsuperscript{1}, Rui Wang\textsuperscript{1}, Zhenglu Yang\textsuperscript{2} \\
\textsuperscript{1} Alibaba Group, China  \\
\textsuperscript{2} School of Computer Science, Nankai University, China  \\
\{\tt guimin.gm, tjf141457, masi.wr\}@alibaba-inc.com\\ \tt yangzl@nankai.edu.cn}

\date{}

\begin{document}
\maketitle
\begin{abstract}

Attention plays a key role in the improvement of sequence-to-sequence-based document summarization models. To obtain a powerful attention helping with reproducing the most salient information and avoiding repetitions, we augment the vanilla attention model from both local and global aspects. We propose an attention refinement unit paired with local variance loss to impose supervision on the attention model at each decoding step, and a global variance loss to optimize the attention distributions of all decoding steps from the global perspective. The performances on the CNN/Daily Mail dataset verify the effectiveness of our methods.


\end{abstract}


\section{Introduction}\label{intro}

Abstractive document summarization \cite{Rush2015,Nallapati2016,Tan2017,fastAbs, CommunicatingAgents} attempts to produce a condensed representation of the most salient information of the document, aspects of which may not appear as parts of the original input text. One popular framework used in abstractive summarization is the sequence-to-sequence model introduced by \citet{seq2seq}. The \textit{attention} mechanism \cite{attention} is proposed to enhance the sequence-to-sequence model by allowing salient features to dynamically come to the forefront as needed to make up for the incapability of memorizing the long input source. 

However, when it comes to longer documents, basic attention mechanism may lead to distraction and fail to attend to the relatively salient parts. Therefore, some works focus on designing various attentions to tackle this issue \cite{Tan2017,bottom}. We follow this line of research and propose an effective attention refinement unit (ARU). Consider the following case. Even with a preliminary idea of which parts of source document should be focused on (attention), sometimes people may still have trouble in deciding which exact part should be emphasized for the next word (the output of the decoder). To make a more correct decision on what to write next, people always adjust the concentrated content by reconsidering the current state of what has been summarized already. Thus, ARU is designed as an update unit based on current decoding state, aiming to retain the attention on salient parts but weaken the attention on irrelevant parts of input. 

The de facto standard attention mechanism is a soft attention that assigns attention weights to all input encoder states, while according to previous work \cite{showAttend,SurprisinglyHardAttention}, a well-trained hard attention on exact one input state is conducive to more accurate results compared to the soft attention. To maintain good performance of hard attention as well as the advantage of end-to-end trainability of soft attention, we introduce a local variance loss to encourage the model to put most of the attention on just a few parts of input states at each decoding step. Additionally, we propose a global variance loss to directly optimize the attention from the global perspective by preventing assigning high weights to the same locations multiple times. 
The global variance loss is somewhat similar with the coverage mechanism \cite{TuLLLL16,See2017}, which is also designed for solving the repetition problem. The coverage mechanism introduces a coverage vector to keep track of previous decisions at each decoding step and adds it into the attention calculation. However, when the high attention on certain position is wrongly assigned during previous timesteps, the coverage mechanism hinders the correct assignment of attention in later steps.

We conduct our experiments on the CNN/Daily Mail dataset and achieve comparable results on ROUGE \cite{rouge} and METEOR \cite{meteor} with the state-of-the-art models. Our model surpasses the strong pointer-generator baseline (w/o coverage) \cite{See2017} on all ROUGE metrics by a large margin. As far as we know, we are the first to introduce explicit loss functions to optimize the attention. More importantly, the idea behind our model is simple but effective. Our proposal could be applied to improve other attention-based models, which we leave these explorations for the future work.

\section{Proposed model}
\subsection{Model Architecture}\label{Baseline Architecture}

We adopt the Pointer-Generator Network (PGN) \cite{See2017} as our baseline model, which augments the standard attention-based seq2seq model with a hybrid pointer network \cite{pointer}. An input document is firstly fed into a Bi-LSTM encoder, then an uni-directional LSTM is used as the decoder to generate the summary word by word. At each decoding step, the attention distribution $a_t$ and the context vector $c_t$ are calculated as follows:
\begin{align}
    e_{ti} = & \, v^T\tanh(W_{h}h_i + W_{s}s_t + b_{attn}) \\
    a_t    = & \, \text{softmax}(e_t) \\
    c_t    = & \, \sum_{i=1} a_{ti} h_i
\end{align}
\noindent where $h_i$ and $s_t$ are the hidden states of the encoder and decoder, respectively. Then, the token-generation softmax layer reads the context vector $c_t$ and current hidden state $s_t$ as inputs to compute the vocabulary distribution. To handle OOVs, we inherit the pointer mechanism to copy rare or unseen words from the input document (refer to \citet{See2017} for more details). 

\begin{figure}[t!]
\includegraphics[width=1\columnwidth]{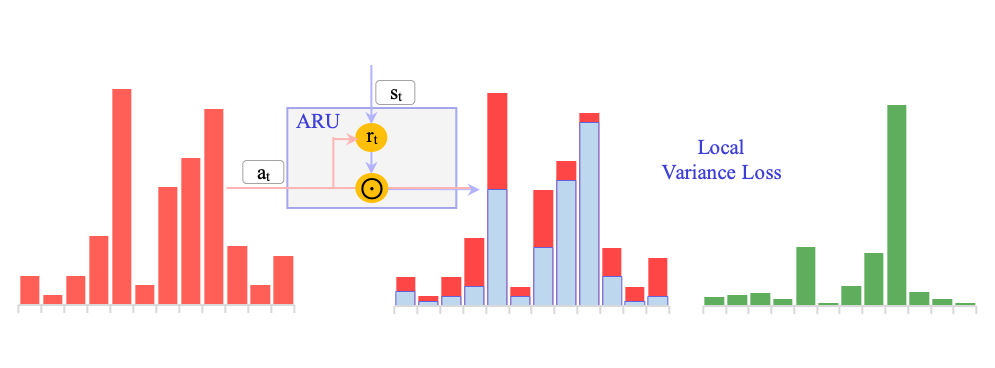}
\caption{The process of attention optimization (better view in color). The original attention distribution (\textcolor[RGB]{255,70,70}{red bar on the left}) is updated by the refinement gate $r_t$ and attention on some irrelevant parts are lowered. Then the updated attention distribution (\textcolor[RGB]{128,183,238}{blue bar in the middle}) is further supervised by a local variance loss and get a final distribution (\textcolor[RGB]{74,171,74}{green bar on the right}).}
\label{fig:model}
\end{figure}

To augment the vanilla attention model, we propose the Attention Refinement Unit (ARU) module to retain the attention on the salient parts while weakening the attention on the irrelevant parts of input. 
As illustrated in Figure \ref{fig:model}, the attention weight distribution $a_t$ at timestep $t$ (the first red histogram) is fed through the ARU module. In the ARU module, current decoding state $s_t$ and attention distribution $a_t$ are combined to calculate a refinement gate $r_t$:
\begin{equation}
    r_t = \sigma (W_{s}^{r}s_t + W_{a}^ra_t + b_r) 
\end{equation}

\noindent where $\sigma$ is the sigmoid activation function, $W_{s}^{r}$, $W_{a}^r$ and $b_r$ are learnable parameters. $r_t$ represents how much degree of the current attention should be updated. Small value of $r_{ti}$ indicates that the content of $i$-th position is not much relevant to current decoding state $s_t$, and the attention on $i$-th position should be weakened to avoid confusing the model. The attention distribution is updated as follows (the symbol $\odot$ means element-wise product):
\begin{equation}
    a_t^r = r_t \odot a_t
\end{equation}


\subsection{Local Variance Loss}
As discussed in section \ref{intro}, the attention model putting most of attention weight on just a few parts of the input tends to achieve good performance. Mathematically, when only a small number of values are large, the shape of the distribution is sharp and the variance of the attention distribution is large. 
Drawing on the concept of variance in mathematics, local variance loss is defined as the reciprocal of its variance expecting the attention model to be able to focus on more salient parts. The standard variance calculation is based on the mean of the distribution. However, as previous work \cite{median,variance} mentioned that the median value is more robust to outliers than the mean value, we use the median value to calculate the variance of the attention distribution. Thus, local variance loss can be calculated as:
\begin{align}
    var(a_t^r) = \frac{1}{|D|}\sum_{i=1}^{|D|}(a^r_{ti} - \hat{a_t^r})^2 \\
    \mathcal{L}_{L} = \frac{1}{T}\sum_{t}^T\frac{1}{var(a_t^r) + \epsilon}
\end{align}

\noindent where $\hat{\cdot}$ is a median operator and $\epsilon$ is utilized to avoid zero in the denominator.

\subsection{Global Variance Loss}
To avoid the model attending to the same parts of the input states repeatedly, we propose another variance loss to adjust the attention distribution globally. Ideally, the same locations should be assigned a relatively high attention weight once at most. Different from the coverage mechanism \cite{See2017, TuLLLL16} tracking attention distributions of previous timesteps, we maintain the sum of attention distributions over all decoder timesteps, denoted as $A$. The $i$-th value of $A$ represents the accumulated attention that the input state at $i$-th position has received throughout the whole decoding process. Without repeated high attention being paid to the same location, the difference between the sum of attention weight and maximum attention weight of $i$-th input state among all timesteps should be small.
Moreover, the whole distribution of the difference over all input positions should have a flat shape. Similar to the definition of local variance loss, the global variance loss is formulated as:
\begin{align}
g_i              &= \sum_t(a_{ti}^r) - \max_t(a_{ti}^r)  \\
\mathcal{L}_{G}  &= \frac{1}{|D|} \sum_{i=1}^{|D|} (g_i - \hat{g})^2
\end{align}
\noindent where $g_i$ represents the difference between the accumulated attention weight and maximum attention weight at $i$-th position.

\subsection{Model Training}
The model is firstly pre-trained to minimize the maximum-likelihood loss, which is widely used in sequence generation tasks. We define $y^* = \{y^*_1, \cdots, y_T^*\}$ as the ground-truth output sequence for a given input sequence $x$, then the loss function is formulated as:
\begin{equation}
    \mathcal{L}_{MLE} = -\frac{1}{T}\sum_{t=1}^T\log(p(y_t^*|x)
\end{equation}
\noindent After converging, the model is further optimized with local variance loss and global variance loss. The mix of loss functions is:
\begin{equation}
    \mathcal{L} = \mathcal{L}_{MLE} + \lambda_1\mathcal{L}_{L} + \lambda_2\mathcal{L}_{G} \label{eq:loss}
\end{equation}
\noindent where $\lambda_1$ and $\lambda_2$ are hyper-parameters.

\setlength{\belowcaptionskip}{-0.13cm} 
\begin{table*}[t]
    \centering
    \begin{tabular}{|l|c|c|c|c|}
    \hline
    \bf Models  & \bf ROUGE-1 & \bf ROUGE-2 & \bf ROUGE-L  & \bf METEOR \\
    \hline 
    \multicolumn{5}{|c|}{PREVIOUS WORKS} \\
    \hline 
    PGN \cite{See2017}         & 36.44 & 15.66 & 33.41 & 16.65 \\
    PGN+Coverage \cite{See2017}        & 39.53 & 17.28 & 36.38 & 18.72\\
    Intra-att.+RL \cite{paulus2018a}    & 39.87 & 15.82 & 36.90 & - \\
    FastAbs+RL  \cite{fastAbs}           & 40.88 & 17.80 & 38.54 & 20.38 \\
    DCA+RL \cite{deepchannel}               & 41.69 & 19.47 & 37.92 & -\\
    \hline
    \multicolumn{5}{|c|}{OUR MODELS} \\
    \hline
    PGN (ours)        & 36.72 & 15.76 & 33.40 & 17.19 \\
    PGN+Coverage (ours)             & 39.75 & 17.42 & 36.36 & 19.73 \\
    PGN+ARU               & 37.41 & 16.01 & 34.05 & 18.03 \\
    +Local variance loss  & 39.45 & 17.26 & 35.99 & 19.02 \\
    +Global variance loss & 40.29 & 17.76 & 36.78 & 19.88 \\
    \hline
    \end{tabular}
    \caption{Performance on CNN/Daily Mail test dataset.}\label{tab:result}
\end{table*}

\section{Experiments}

\subsection{Preliminaries}
\paragraph{Dataset and Metrics.} We conduct our model on the large-scale dataset CNN/Daily Mail \cite{cnndm,Nallapati2016}, which is widely used in the task of abstractive document summarization with multi-sentences summaries. We use the scripts provided by \citet{See2017} to obtain the non-anonymized version of the dataset without preprocessing to replace named entities. The dataset contains 287,226 training pairs, 13,368 validation pairs and 11,490 test pairs in total. We use the full-length ROUGE F1\footnote{We use the official package pyrouge \url{https://pypi.org/project/pyrouge/}} and METEOR\footnote{\url{http://www.cs.cmu.edu/~alavie/METEOR/}} as our main evaluation metrics.

\paragraph{Implementation Details.} The data preprocessing is the same as PGN \cite{See2017}, and we randomly initialize the word embeddings. The hidden states of the encoder and the decoder are both 256-dimensional and the embedding size is also 256. Adagrad with learning rate 0.15 and an accumulator with initial value 0.1 are used to train the model. We conduct experiments on a single Tesla P100 GPU with a batch size of 64 and it takes about 50000 iterations for pre-training and 10000 iterations for fine-tuning. Beam search size is set to 4 and trigram avoidance \cite{paulus2018a} is used to avoid trigram-level repetition. 
Tuned on validation set, $\lambda_1$ and $\lambda_2$ in the loss function (Equation. \ref{eq:loss}) is set to 0.3 and 0.1, respectively.

\subsection{Automatic Evaluation Result}
As shown in Table \ref{tab:result} (the performance of other models is collected from their papers), our model exceeds the PGN baseline by 3.85, 2.1 and 3.37 in terms of R-1, R-2 and R-L respectively and receives over 3.23 point boost on METEOR. FastAbs \cite{fastAbs} regards ROUGE scores as reward signals with reinforcement learning, which brings a great performance gain. DCA \cite{CommunicatingAgents} proposes deep communicating agents with reinforcement setting and achieves the best results on CNN/Daily Mail. Although our experimental results have not outperformed the state-of-the-art models, our model has a much simpler structure with fewer parameters. Besides, these simple methods do yield a boost in performance compared with PGN baseline and may be applied on other models with attention mechanism. 

\begin{figure}
\setlength{\belowcaptionskip}{-0.5cm} 
\centering\includegraphics[width=7.2cm]{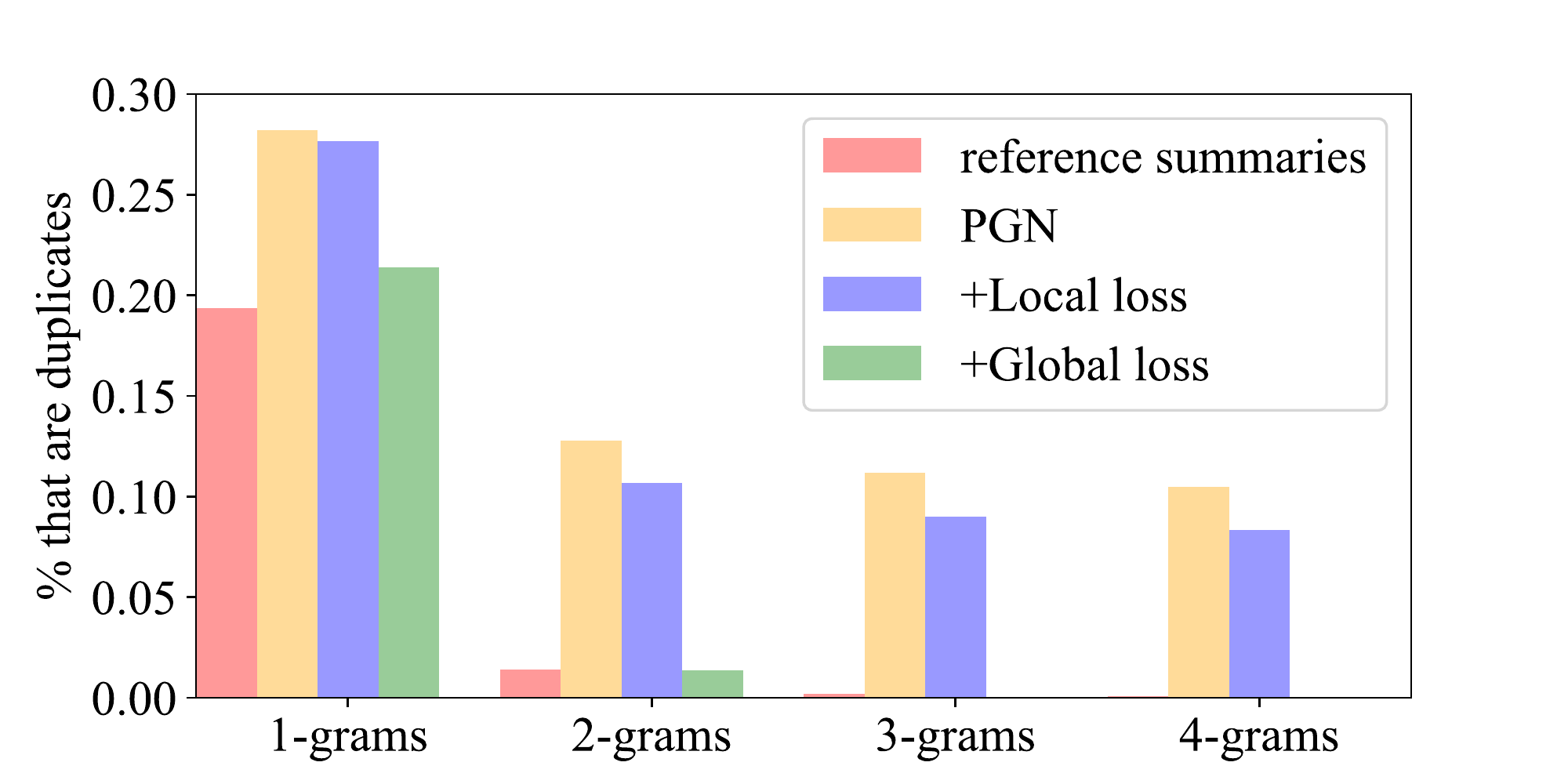}
\caption{With global variance loss, our model (\textcolor[RGB]{65,171,65}{green bar}) can avoid repetitions and achieve comparable percentage of duplicates with reference summaries.}
\label{fig:duplication}
\end{figure}

We further evaluate how these optimization approaches work. 
The results at the bottom of Table \ref{tab:result} verify the effectiveness of our proposed methods. The ARU module has achieved a gain of 0.97 ROUGE-1, 0.35 ROUGE-2, and 0.64 ROUGE-L points; the local variance loss boosts the model by 3.01 ROUGE-1, 1.6 ROUGE-2, and 2.58 ROUGE-L. As shown in Figure \ref{fig:duplication}, the global variance loss helps with eliminating n-gram repetitions, which verifies its effectiveness.

\subsection{Human Evaluation and Case Study} 

We also conduct human evaluation on the generated summaries. Similar to the previous work \cite{fastAbs, Nallapati2017}, we randomly select 100 samples from the test set of CNN/Daily Mail dataset and ask 3 human testers to measure \emph{relevance} and \emph{readability} of each summary. Relevance is based on how much salient information does the summary contain, and readability is based on how fluent and grammatical the summary is. 
Given an article, different people may have different understandings of the main content of the article, the ideal
situation is that more than one reference is paired with the articles.
However, most of summarization datasets contain the pairs of article with a single reference summary due to the cost of annotating multi-references. Since we use the reference summaries as target sequences to train the model and assume that they are the gold standard, we give both articles and reference summaries to the annotator to score the generated summaries. In other words, we compare the generated summaries against the reference ones and the original article to obtain the (relative) scores in Table 3. Each perspective is assessed with a score from 1 (worst) to 5 (best). The result in Table \ref{tab:human} demonstrate that our model performs better under both criteria w.r.t. \citet{See2017}. Additionally, we show the example of summaries generated by our model and baseline model in Table \ref{fig:example}. As can be seen from the table, PGN suffers from repetition and fails to obtain the salient information. Though with coverage mechanism solving saliency and repetition problem, it generates many trivial facts. 
With ARU, the model successfully concentrates on the salient information, however, it also suffers from serious repetition problem. Further optimized by the variance loss, our model can avoid repetition and generate summary with salient information. Besides, our generated summary contains fewer trivial facts compared to the PGN+Coverage model.

\begin{table}[t]
    \centering
    \begin{tabular}{|l|c|c|}
    \hline
    \bf Models   & \bf Relevance & \bf Readability \\
    \hline
    Reference    & 5.00  & 5.00 \\
    PGN         & 2.27  & 4.30 \\
    PGN+Coverage & 2.46  & 4.88 \\
    \hline
    Our model    & 2.74  & 4.92 \\
    \hline
    \end{tabular}
    \caption{Human Evaluation: pairwise comparison
between our final model and PGN model.}\label{tab:human}
\end{table}

\begin{table*}[t]
\small
\begin{tabular}{|p{2.01\columnwidth}|}
\hline 
\indent {\bf Article}:
poundland has been been forced to pull decorative plastic easter eggs from their shelves over fears children may choke - because they look like cadbury mini eggs . \textbf{trading standards officials in buckinghamshire and surrey raised the alarm} over the chinese made decorations , as \textbf{they were ` likely to contravene food imitation safety rules ' }. the eggs have now been withdrawn nationwide ahead of the easter break . scroll down for video . poundland has been been forced to pull decorative plastic easter eggs from their shelves over fears they may choke - because they look like cadbury mini eggs -lrb- pictured is the poundland version -rrb- . \textbf{the eggs bear a striking similarity to the sugar-coated chocolate treats} with a brown ` speckle ' designed to make it look like a quail 's egg -lrb- cadbury mini eggs are pictured -rrb- . \dots ` parents should also be wary of similar products being offered for sale over the easter period at other stores or online . '\\
\hline
\indent {\bf Reference Summary}:
Trading standards officials in buckinghamshire and surrey raised alarm. Officers said they were `likely to contravene food imitation safety rules'. The eggs bear a striking similarity to the sugar-coated chocolate treats. \\
\hline
\indent {\bf PGN}: 
Poundland has been forced to pull decorative plastic easter eggs from their shelves over fears children may choke - because they look like cadbury mini eggs. The eggs have now been withdrawn nationwide ahead of the easter break. \emph{\textbf{\textcolor[RGB]{255,67,67}{The eggs have now been withdrawn nationwide ahead of the easter break.}}} \\
\hline
\indent{\bf PGN+Coverage}: \textbf{\textcolor[RGB]{105,105,255}{Trading standards officials in buckinghamshire and surrey raised the alarm}} over the chinese made decorations , as \textbf{\textcolor[RGB]{105,105,255}{they were ` likely to contravene food imitation safety rules '}} the eggs have now been withdrawn nationwide ahead of the easter break . \textbf{\textcolor[RGB]{105,105,255}{the eggs bear a striking similarity to the sugar-coated chocolate treats}} with a brown ` speckle ' designed to make it look like a quail 's egg .\\
\hline
\indent {\bf + ARU}:
\textbf{\textcolor[RGB]{105,105,255}{Eggs bear a striking similarity to the sugar-coated chocolate treats}} with a brown `speckle' designed to make it look like a quail's egg. \emph{\textbf{\textcolor[RGB]{255,67,67}{The eggs bear a striking similarity to the sugar-coated chocolate treats with a brown `speckle' designed to make it look like a quail's egg.}}} \\
\hline
\indent {\bf + Variance loss}:
\textbf{\textcolor[RGB]{105,105,255}{Trading standards officials in buckinghamshire and surrey raised the alarm}} over the chinese made decorations, as \textbf{\textcolor[RGB]{105,105,255}{they were `likely to contravene food imitation safety rules'}}. The eggs have now been withdrawn nationwide ahead of the easter break. \textbf{\textcolor[RGB]{105,105,255}{The eggs bear a striking similarity to the sugar-coated chocolate treats}} with a brown `speckle'. \\
\hline
\end{tabular}
\caption{The \textbf{bold} words in \textit{article} are salient parts contained in \textit{reference summary}. The \textbf{\textcolor[RGB]{105,105,255}{blue}} words in generated summaries are salient information and the \emph{\textbf{\textcolor[RGB]{255,67,67}{red}}} words are repetition.}
\label{fig:example}
\end{table*}


\section{Related Work}
The exploration on document summarization can be broadly divided into extractive and abstractive summarization. The extractive methods \cite{Nallapati2017,SWAP-NET,deepchannel} select salient sentences from original document as a summary. In contrast, abstractive summarization \cite{Rush2015, Nallapati2016,See2017,fastAbs} generates summaries word-by-word after digesting the main content of the document. Out-of-vocabulary(OOV), repetition, and saliency are three conspicuous problems need to be well solved in abstractive document summarization. Some works \cite{Nallapati2016,See2017,paulus2018a} handle the OOV problem by introducing the pointer network. \citet{See2017} introduces a coverage mechanism, which is a variant of the coverage vector \cite{TuLLLL16} from Neural Machine Translation, to eliminate repetitions. However, there are just a few studies on saliency problem \cite{Tan2017,deepchannel,bottom}. To obtain more salient information, \citet{ChenIJCAI2016} proposes a new attention mechanism to distract them in the decoding step to better grasp the overall meaning of input documents. We optimize attention using an attention refinement unit under the novel variance loss supervision. As far as we know, we are the first to propose explicit losses to refine the attention model in abstractive document summarization tasks. 
Recently many models \cite{paulus2018a,CommunicatingAgents,fastAbs,ScoreSelectSentences,closedBooktraining} have emerged taking advantage of reinforcement learning (RL) to solve the discrepancy issue in seq2seq model and have yielded the state-of-the-art performance.


\section{Conclusion}
In this paper, we propose simple but effective methods to optimize the vanilla attention mechanism in abstarctive document summarization. The results on CNN/Daily Mail dataset demonstrate the effectiveness of our methods. We argue that these simple methods are also adaptable to other summarization models with attention. Further exploration on this and combination with other approaches like RL remains as our future exploration. Besides, we will also conduct experiments on several other current summarization datasets like New York Times (NYT) \cite{paulus2018a} and Newsroom \cite{newsroom}.

\bibliography{emnlp-ijcnlp-2019}
\bibliographystyle{acl_natbib}

\end{document}